# A MapReduce based distributed SVM algorithm for binary classification

Ferhat Özgür Çatak[1], Mehmet Erdal Balaban[2]

[1]National Research Institute of Electronics and Cryptology, TUBITAK, Turkey,
Tel: 0-262-6481070, e-mail: ozgur.catak@tubitak.gov.tr

[2]Faculty of Business Administration, Quantitative Methods, Istanbul University, Turkey,
Tel: 0-212-4400000, e-mail: balaban@gmail.com

**Abstract**

Although Support Vector Machine (SVM) algorithm has a high generalization property to classify for unseen examples after training phase and it has small loss value, the algorithm is not suitable for real-life classification and regression problems. SVMs cannot solve hundreds of thousands examples in training dataset. In previous studies on distributed machine learning algorithms, SVM is trained over a costly and preconfigured computer environment. In this research, we present a MapReduce based distributed parallel SVM training algorithm for binary classification problems. This work shows how to distribute optimization problem over cloud computing systems with MapReduce technique. In the second step of this work, we used statistical learning theory to find the predictive hypothesis that minimize our empirical risks from hypothesis spaces that created with reduce function of MapReduce. The results of this research are important for training of big datasets for SVM algorithm based classification problems. We provided that iterative training of split dataset with MapReduce technique; accuracy of the classifier function will converge to global optimal classifier function's accuracy in finite iteration size. The algorithm performance was measured on samples from letter recognition and pen-based recognition of handwritten digits dataset.

## 1. Introduction

Most of machine learning algorithms have problems with computational complexity of training phase with large scale learning datasets. Applications of classification algorithms for large scale dataset are computationally expensive to process. The computation time and storage space of Support Vector Machine (SVM) algorithm are very largely determined by large scale kernel matrix [1]. Computational complexity and the computation time are always limiting factor for machine learning in practice. In order to overcome this complexity problem, researchers developed some techniques; feature selection, feature extraction and distributed computing.

Feature selection methods are used for machine learning model construction with reduced number of features. Feature selection is a basic approach for reducing feature vector size [2]. A new combination of feature subset is obtained with various algorithms such as information gain [3], correlation based feature selection [4], Gini index [5] and t-statistics. Feature selection methods solve two main problems. The first solution is reducing the number of the feature set in the training set to effectively use of computing resources like memory and CPU and second solution is to remove noisy features from the dataset in order to improve the classification algorithm performance [6].

Feature extraction methods are used to achieve the curse of dimensionality that refers to the problems as the dimensionality increases. In this approach, high dimensional feature space is transformed into low dimensional feature space. There are several feature extraction algorithms such as Principal Component Analysis (PCA) [7], Singular Value Decomposition (SVD) [8], Independent Component Analysis (ICA) [9].

The last solution to overcome the large amount of memory and computation power requirements for training large scale dataset is chunking or distributed computing [10]. Graf et al. [11] proposed the cascade SVM to overcome very large scale classification problems. In this method, dataset is split into n parts in feature space. Non-support vectors of each sub dataset are filtered and only support vectors are transmitted. The margin optimization process uses only combined sub dataset to find out the support vectors. Collobert et al. [12] proposed a new parallel SVM training and classification algorithm that each subset of a dataset is trained with SVM and then the classifiers are combined into a final single classifier function. Lu et al. [13] proposed strongly connected network based distributed support vector machine algorithm. In this method, dataset is split into k roughly equal part for each computer in a network then, support vectors are exchanged among these computers. Ruping et al. [14] proposed a novel incremental learning with SVM algorithm. Syed et al. [15] proposed another incremental learning method. In this method, a fusion center collects all support vectors from distributed computers. Caragea et al. [16] used previous method. In this algorithm, fusion center iteratively sends support vectors back to computers. Sun et al. [17] proposed a novel method for parallelized SVM based on MapReduce technique. This method is based on the cascade SVM model. Their approach is based on iterative MapReduce model Twister which is different from our implementation of Hadoop based MapReduce. Their method is same with cascade SVM model. They use only support vectors of a sub dataset to find an optimal classifier function. Another difference from our approach is that they apply feature selection with correlation coefficient method for reducing number of feature in datasets before training the SVM to improve the training time.

In our previous research [18], we developed a novel approach for MapReduce based SVM training for binary classification problem. We used some UCI dataset to show generalization property of our algorithm.

In this paper, we propose a novel approach and formal analysis of the models that generated with the MapReduce based binary SVM training method. We distribute whole training dataset over data nodes of cloud computing system. At each node, subset of training dataset is used for training to find out a binary classifier function. The algorithm collects support vectors (SVs) from every node in cloud computing system, and then merges all SVs to save as global SVs. Our algorithm is analyzed with letter recognition [19] and pen-based recognition of handwritten digits [20] dataset with Hadoop streaming using MrJob python library. Our algorithm is built on the LibSVM and implemented using the Hadoop implementation of MapReduce.

The organization of this article is as follows. In the next section, we will provide an overview to SVM formulations. In Section 3, we present the MapReduce pattern in detail. Section 4 explains the system model with our implementation of MapReduce pattern for the SVM training. In section 5, convergence of our algorithm is explained. In section 6, simulation results with letter recognition and pen-based recognition of handwritten digits datasets are shown. Thereafter, we will give concluding remarks in Section 7.

## 2. Support Vector Machine

In machine learning field, SVM is a supervised learning algorithm for classification and regression problems depending of the type of output. SVM uses statistical learning theory to maximize generalization property of generated classifier model. SVM avoids over fitting to the training dataset. Statistical learning theory generalizes the quality of fitting the training data (empirical error). Empirical risk is $R = \frac{1}{n}\sum_{i=1}^{n} \ell(f_\theta(x_i), y_i)$ which is the average loss $\ell$ of the chosen estimator over the training set $\{(x_i, y_i)\}$ [21]. SVM use a set of training data and predicts, for each given input, which of two possible class $\{-1,1\}$. As shown in Figure 1, the hyperplane is defined by $w^T x + b = 0$, where $w \in R^n$ is a orthogonal to the hyperplane and $b \in R^n$ is the bias. Giving some training data $\mathcal{D}$, a set of point of the form

$$\mathcal{D} = \{(x_i, y_i) \mid x_i \in R^m, y_i \in \{-1,1\}\}_{i=1}^{n} \qquad (1)$$

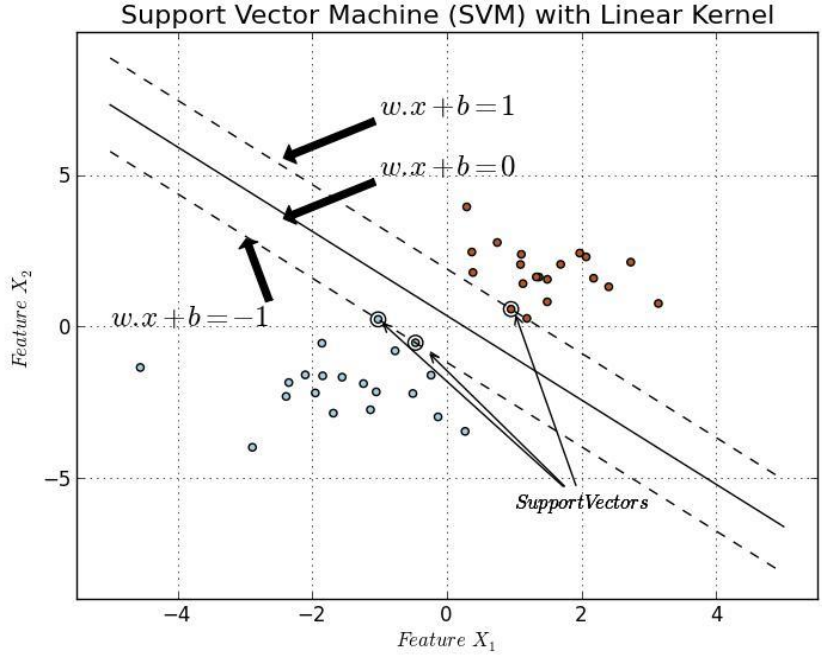

**Figure 1** Classification of an SVM with Maximum-margin hyper plane trained with samples from two classes.

where $x_i$ is a m-dimensional real vector, $y_i$ is the class of input vector $x_i$ either -1 or 1. SVMs aim to search a hyper plane that maximizes the margin between the two classes of samples in $\mathcal{D}$ with the smallest empirical risk [22]. For the generalization property of SVM, two parallel hyperplanes are defined such that $w^T x + b = 1$ and $w^T x + b = -1$. One can simplify these two functions into new one.

$$y_i(\boldsymbol{w}^T \boldsymbol{x}_i + b) \geq 1 \qquad (2)$$

SVM aims to maximize distance between these two hyperplanes. One can calculate the distance between these two hyperplanes with $\frac{1}{\|w\|}$. The training of SVM for the non-separable case is solved using quadratic optimization problem that shown in Equation 3.

$$minimize : P(\mathbf{w}, b, \xi) = \frac{1}{2}\|\mathbf{w}\|^2 + C\sum_{i=1}^{m}\xi_i$$

$$subject\ to : y_i(\langle \mathbf{w}, \phi(\mathbf{x}_i)\rangle + b) \geq 1 - \xi_i$$
$$\xi_i \geq 0$$

(3)

for $i = 1, \ldots, m$, where $\xi_i$ are slack variables and C is the cost variable of each slack. C is a control parameter for the margin maximization and empirical risk minimization. The decision function of SVM is $f(x) = w^T \phi(x) + b$ where the w and b are calculated by the optimization problem P in Equation (3). By using Lagrange multipliers, the optimization problem P in Equation (3) can be expressed as

$$min: F(\alpha) = \frac{1}{2}\boldsymbol{\alpha}^T \mathbf{Q}\, \boldsymbol{\alpha}^T - \boldsymbol{\alpha}^T \mathbf{1}$$
$$subject\ to : 0 \leq \boldsymbol{\alpha} \leq C$$
$$y^T \boldsymbol{\alpha} = 0$$

(4)

where $[Q]_{ij} = y_i y_j \phi^T(x_i)\phi(x_j)$ is the Lagrangian multiplier variable. It is not needed to know function $\phi$, but it is necessary to know how to compute the modified inner product which will be called as kernel function represented as $K(x_i, x_j) = \phi^T(x_i)\phi(x_j)$. Thus, $[Q]_{ij} = y_i y_j K(x_i, x_j)$ [23].

## 3. Map Reduce Model

MapReduce is a programming model derived from the map and reduces function combination from functional programming. MapReduce model widely used to run parallel applications for large scale datasets processing. MapReduce uses key/value pair data type in map and reduce functions. [24]. Overview of MapReduce system is show in Figure 2.

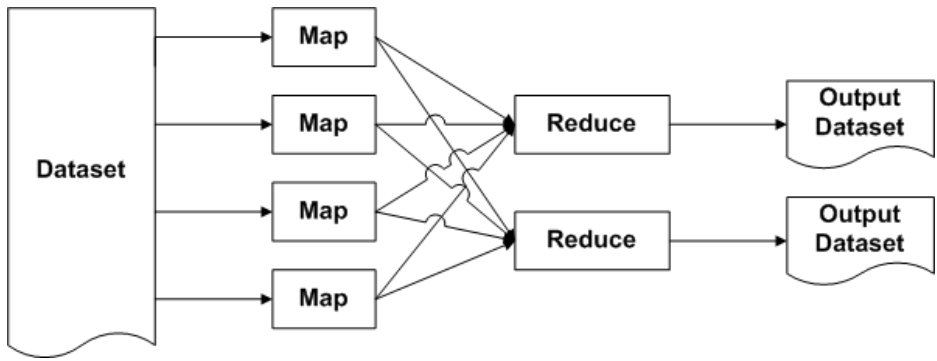

**Figure 2** Overview of MapReduce System

MapReduce pattern is divided into two functions which are map and reduce. These two functions are separated by a shuffle step of the intermediate key/value data. The MapReduce framework executes those functions in parallel manner over any number of computers [25]. Simply, a MapReduce job executes three basic operations on a dataset distributed across many shared-nothing cluster nodes. The first task is Map function that processes in parallel manner by each node without transferring any data with other notes. In next operation, processed data by Map function is repartitioned across all nodes of the cluster. Lastly, Reduce task is executed in parallel manner by each node with partitioned data.

A file in the distributed file system (DFS) is split into multiple chunks and each chunk is stored on different data nodes. The input of a map function is a key/value pair from input chunks of dataset and it creates an output in list of key/value pairs:

$$map(key_1, value_1) \Rightarrow list(key_2, value_2)$$

A reduce function takes a key value and its value list as input. Then, reduce function generates a list of new values as output:

$$reduce(key_2, list(value_2)) \Rightarrow list(value_3)$$

## 4. System Model

The cloud computing based binary class support vector machine algorithm works as follows. The training set of the algorithm is split into subsets. Each node within a cloud computing system classifies sub dataset locally via SVM algorithm and gets α values (i.e. support vectors (SVs)), and then passes the calculated SVs to global SVs to merge them. In Map stage of MapReduce job, the subset of training set is combined with global support vectors. In Reduce step, the merged subset of training data is evaluated. The resulting new support vectors are combined with the global support vectors in Reduce step. The algorithm can be explained as follows. First, each node in a cloud computing system reads the global support vectors set, then merges global SVs set with subsets of local training dataset and classifies using SVM algorithm. Finally, all the computed SVs set in cloud nodes are merged. Thus, algorithm saves global SVs set with new ones. Our algorithm consists of the following steps. We showed our terminology at Table 1.

**Table 1**: The notation we used in our work.

| Notation | Description |
| --- | --- |
| $t$ | Iteration number |
| $L$ | Number of computers (or MapReduce function size) |
| $h^t$ | Best hypothesis at iteration $t$ |
| $D_l$ | Sub data set at computer $l$ |
| $SV_l$ | Support vectors at computer $l$ |
| $SV_{Global}$ | Global support vector |

1. As initialization, the global support vector set as $t = 0, SV^t = \emptyset$
2. t = t + 1;
3. For any computer in $l, l = 1, \ldots, L$ reads global SVs and merge them with subset of training data.
4. Train SVM algorithm with merged new dataset
5. Find out support vectors
6. After all computers in cloud system complete their training phase, merge all calculated SVs and save the result to the global SVs
7. If $h^t = h^{t-1}$ stop, otherwise go to step 2

---

**Algorithm 1** Map Function of Binary SVM Algorithm

---

$SV_{Global} = \emptyset$ // *Empty global support vector set*
**while** $h^t \neq h^{t-1}$
  **for** $l \in L$ **do** // *For each subset loop*
    $D_l^t \leftarrow D_l^t \cup SV_{Global}^t$
  **end for**
**end while**

---

**Algorithm 2** Reduce Function of Binary SVM Algorithm

---

**while** $h^t \neq h^{t-1}$ **do**
  **for** $l \in L$
// *Train merged Dataset to obtain Support*
// *Vectors and Binary-Class Hypothesis*
    $SV_l, h^t \leftarrow binarySvm(D_l)$
  **end for**
  **for** $l \in L$
    $SV_{Global} \leftarrow SV_{Global} \cup SV_l$
  **end for**
**end while**

---

Pseudo code of our algorithm's Map and Reduce function are given in Algorithm 1 and Algorithm 2.

For training SVM classifier functions, we used LibSVM with various kernels. Appropriate parameters C and γ values were found by cross validation test. We used 10-fold cross validation method. All system is implemented with Hadoop and streaming Python package mrjob library.

## 5. Convergence of The Algorithm with Statistical Learning Theory

Let $\mathcal{S}$ denotes a subset of training dataset $\mathcal{D}$, $F(\mathcal{S})$ is the optimal classifier function over dataset $\mathcal{S}$, $h^*$ is the global optimal hypothesis for which has a minimal empirical risk $R_{emp}(h)$ over dataset $\mathcal{D}$, $Y_\mathcal{S}$ is the vector space of all possible outputs over sub dataset $\mathcal{S}$. Our algorithm's aim is to find a classifier function $f: X \rightarrow Y$ such that $f(x) \sim y$. Let $\mathcal{H}$ be hypothesis space of functions $f: X \rightarrow Y$. Our algorithm starts with $SV_{Global}^0 = \emptyset$, and generates a non-increasing sequence of positive set of vectors $SV_{Global}^t$, where $SV_{Global}^t$ is the vector of support vector at the t.th iteration. We used hinge loss for testing our models trained with our algorithm. Hinge loss works well for its purposes in SVM as a classifier, since the more you violate the margin, the higher the penalty is [26]. The hinge loss function is the following:

$$l(f(x), y) = max\{0, 1 - y.f(x)\}y_i \quad (5)$$

Empirical risk can be computed with an approximation:

$$R_{emp}(h) = \frac{1}{n}\sum_{i=1}^{n}\left(l(h(x_i), y_i)\right) \quad (6)$$

According to the empirical risk minimization principle the binary class learning algorithm should choose a hypothesis $\hat{h}$ in hypothesis space $\mathcal{H}$ which minimizes the empirical risk:

$$\hat{h} = arg\max_{h \in \mathcal{H}} R_{emp}(h) \quad (7)$$

A hypothesis is found in every cloud node. Let X be a subset of training data at cloud node i where $X \in R^{mxn}$, $SV_{Global}^t$ is the vector of support vector at the t.th iteration, $h^{t,i}$ is hypothesis at node i with iteration t.

Algorithm's stop point is reached when the hypothesis' empirical risk is same with previous iteration. That is:

$$R_{emp}(h^t) = R_{emp}(h^{t-1}) \qquad (11)$$

Lemma: Accuracy of the classifier function of our algorithm at iteration t is always greater or equal to the maximum accuracy of the classifier function at iteration t − 1. That is

$$R_{emp}(h^t) \leq arg \min_{h \in \mathcal{H}^{t-1}} R_{emp}(h) \qquad (12)$$

Proof: Without loss of generality, iterated MapReduce binary class SVM monotonically converges to an optimum classifier.

$$SV_{Global}^t = SV_{Global}^{t-1} \cup \{ SV_i^{t-1} \mid i = 1, \dots n \} \qquad (13)$$

where n is the dataset split size (or cloud node size). Then, training set for SVM algorithm at node $i$ is

$$d = X \cup SV_{Global}^t \qquad (14)$$

Adding more samples cannot decrease the optimal value. Generalization accuracy of the sub problem in each node monotonically increases in each iteration step.

## 6. Simulation Results

Our experimental datasets are real handwriting data. The first dataset, the pen-based recognition of handwriting digit dataset [20] contains 250 samples from 44 different writers. All input features are numerical. The classification feature of the dataset is in the range from 0 to 9. The second dataset is letter recognition dataset which contains capital letters with 20 different fonts.

Linear kernels were used with optimal parameters (C, γ). Parameters were estimated by cross-validation method. In our experiments, datasets are randomly partitioned into 10 sub dataset approximately equal-size parts. We ensured that all sub datasets are balanced and classes are uniformly distributed. We fit the classifier function with 90% of original dataset and then using this classifier function we predict the class of 10% remaining test dataset. The cross-validation process is repeated 10 times, with each part is used once as the test samples. We sum the errors on all 10 parts together to calculate the overall error.

## 6.1. Computation Time Comparison Between SVM and MapReduce Based SVM

In our experiments, we compared the single node SVM training algorithm with MapReduce based SVM training algorithm. We used the single node training model as the baseline to find the speedup. Calculation of the speedup is computation time with MapReduce divided by the single node training model computation time. We showed the different node size computation results in Table 2 and Table 3.

**Table 2**: Letter recognition dataset SVM training speedup using MapReduce with different node size.

| Num. of MapReduce Job | Speedup |
|---|---|
| 1 | 1.00 |
| 2 | 3.39 |
| 4 | 4.45 |
| 6 | 4.76 |
| 8 | 5.97 |
| 10 | 6.42 |

**Table 3**: The pen-based recognition of handwriting digit dataset SVM training speedup using MapReduce with different node size.

| Num. of MapReduce Job | Speedup |
|---|---|
| 1 | 1.00 |
| 2 | 2.72 |
| 4 | 4.39 |
| 6 | 4.56 |
| 8 | 6.46 |
| 10 | 7.78 |

The speedups in both data sets are from 6x to 7x. The speedup shown in Table 1 and Table 2 is the average of fifty runs.

## 6.2. Results with MapReduce Based SVM

Figure 3 shows the average accuracy of the test error for each dataset. The figure shows the improvement in MapReduce based SVM at each iteration and stability on large datasets. Figure 4 shows the average number of SVs for each dataset. The figure shows the stability of the number of SVs with MapReduce based SVM at each iteration.

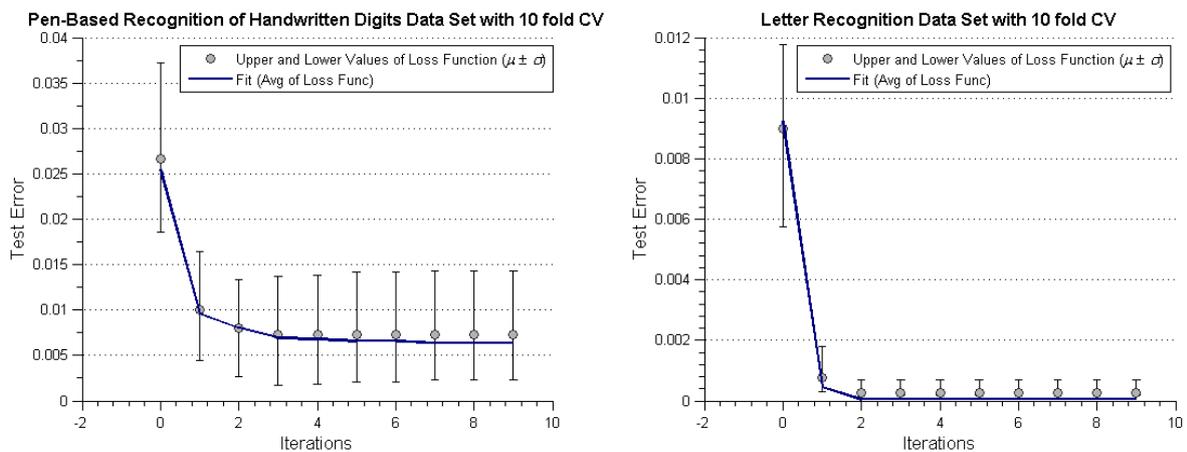

**Figure 3** Hinge loss values over iterations with two datasets.

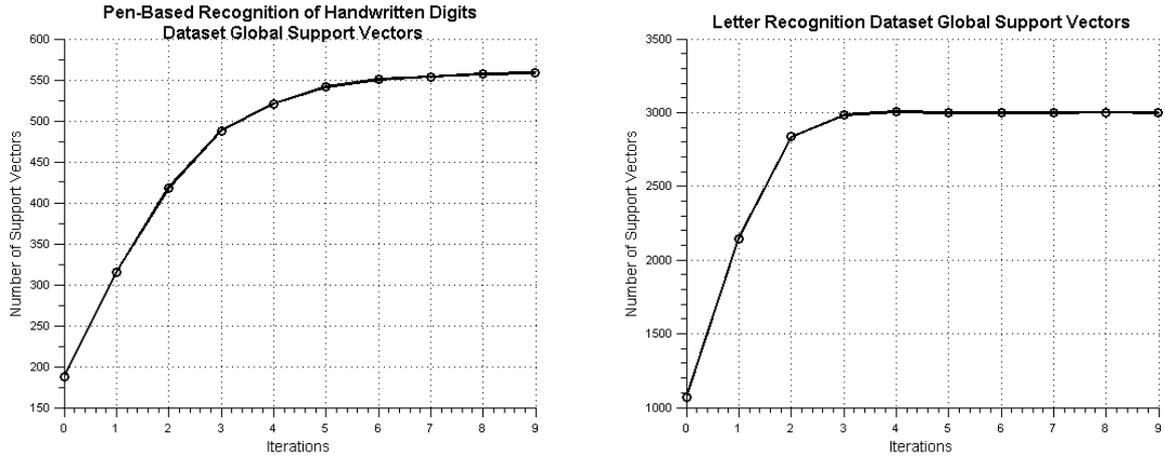

**Figure 4** Support vector sizes over iterations with two datasets.

To analyze our algorithm, we randomly distributed all the training data to a cloud computing system with 10 computers with pseudo distributed Hadoop. We developed python script for distributed support vector machine algorithm with scikit, scipy, numpy, mrjob, matplotlib and libsvm. Dataset prediction accuracies with iterations are shown in Table 4 and Table 5.

**Table 4**: Average, max. and min. value of hinge loss for the pen-based recognition of handwriting digit dataset with 10 fold cross validation.

| Iter. No | Loss($\mu$) | Loss($\mu + \sigma$) | Loss($\mu - \sigma$) |
|---:|---:|---:|---:|
| 1 | 0.02550 | 0.03605 | 0.01736 |
| 2 | 0.00961 | 0.01602 | 0.00401 |
| 3 | 0.00801 | 0.01335 | 0.00267 |
| 4 | 0.00694 | 0.01335 | 0.00134 |
| 5 | 0.00681 | 0.01335 | 0.00134 |
| 6 | 0.00654 | 0.01335 | 0.00134 |
| 7 | 0.00654 | 0.01335 | 0.00134 |
| 8 | 0.00641 | 0.01335 | 0.00134 |
| 9 | 0.00641 | 0.01335 | 0.00134 |
| 10 | 0.00641 | 0.01335 | 0.00134 |

**Table 5**: Average, max. and min. value of hinge loss for the letter recognition dataset with 10 fold cross validation.

| Iter. No | Loss($\mu$) | Loss($\mu + \sigma$) | Loss($\mu - \sigma$) |
|---|---|---|---|
| 1 | 0.00925 | 0.01201 | 0.00600 |
| 2 | 0.00045 | 0.00150 | 0.00000 |
| 3 | 0.00005 | 0.00050 | 0.00000 |
| 4 | 0.00005 | 0.00050 | 0.00000 |
| 5 | 0.00005 | 0.00050 | 0.00000 |
| 6 | 0.00005 | 0.00050 | 0.00000 |
| 7 | 0.00005 | 0.00050 | 0.00000 |
| 8 | 0.00005 | 0.00050 | 0.00000 |
| 9 | 0.00005 | 0.00050 | 0.00000 |
| 10 | 0.00005 | 0.00050 | 0.00000 |

Total numbers of SVs are shown in Table 6. When iteration size becomes 5, test accuracy values of all datasets reach to the highest values. That's the smallest value of the hinge loss of empirical error. If the iteration size is increased, the value of test accuracy falls into a steady state. The value of test accuracy is not changed for large enough number of iteration size.

**Table 6**: Average support vectors size for pen-based recognition of handwriting digit and letter recognition dataset with 10 fold cross validation.

| Iter. No | Pen digit. | Letter recognition |
|---|---|---|
| 1 | 1068.7 | 186.9 |
| 2 | 2147.6 | 314.9 |
| 3 | 2837.7 | 418.2 |
| 4 | 2981.1 | 487.6 |
| 5 | 3003.8 | 520.4 |
| 6 | 2995.8 | 541.0 |
| 7 | 2996.7 | 550.1 |
| 8 | 2996.5 | 553.8 |
| 9 | 2997.5 | 556.9 |
| 10 | 3001.0 | 558.2 |

# 7. Conclusion

In this article, we proposed a new MapReduce based distributed and parallel binary class SVM classification implementation in cloud computing systems with MapReduce model. We showed the generalization property of our algorithm with 10-fold cross validation method. The results of the empirical analyses performed show that our algorithm reaches a steady state condition approximately in 5 iterations. Our research differs from the previous distributed or parallel works mainly in two points. Firstly, we used full datasets for training SVM algorithm. And, the second one, we used binary class classification to obtain classifier function using structural risk minimization property of statistical learning theory. Our approach is simple to implement in another development environments like Java, Matlab etc.

The big data term is used quite frequently nowadays. Most of the datasets used in machine learning fields such as human genome, social networks, and complex physics simulation can be classified as big data. The results of this research are important for training of big datasets for SVM algorithm based classification problems. In the future works, we are planning to use this algorithm in multi-class classification problems with iterative approach of MapReduce with Twister.

## References

[1] Bakır GH, Planck M, Bottou L, Weston J. Breaking SVM complexity with cross training. In: 17 th Neural Information Processing Systems Conference; 5-8 December 2005; Vancouver, Canada: MIT Press. pp. 81-88.

[2] Weston J, Mukherjee S, Chapelle O, Pontil M, Poggio T, Vapnik V. Feature selection for SVMs. In: 14th Annual Neural Information Processing Systems Conference (NIPS); 1-4 December 2000; Denver, USA. 2000 MIT Press. pp. 668-674.

[3] Kullback S, Leibler RA. On Information and Sufficiency. The Annals of Mathematical Statistics 1951; 22: 79–86.

[4] Hall MA. Correlation-based feature selection for machine learning. PhD, The University of Waikato, Hamilton, New Zeland, 1999.

[5] Raileanu LE, Stoffel K. Theoretical comparison between the Gini Index and Information Gain criteria. Annals of Mathematics and Artifcial Intelligence 2004; 41: 77-93.

[6] Mladenić D, Brank J, Grobelnik M, Milic-Frayling N. Feature selection using linear classifier weights: interaction with classification models. In: 27th Annual International ACM SIGIR Conference; 25-29 July 2004; New York, USA: ACM. pp 234-241.

[7] Jolliffe IT. Principal Component Analysis. New York, NY, USA: Springer, 2002.

[8] Golub GH, Reinsch C. Singular value decomposition and least squares solutions. Numerische Mathematik 1970; 10: 403-420.

[9] Common P. Independent Component Analysis, a new concept? Signal Processing 1994 ; 36: 287-314.

[10] Vapnik V. The nature of statistical learning theory. New York, NY, USA: Springer, 1995.

[11] Graf HP, Cosatto E, Bottou L, Durdanovic I, Vapnik V. Parallel support vector machines: The cascade svm. In: Advances in Neural Information Processing Systems; 5-8 December 2005; Vancouver, Canada: MIT Press. pp. 521-528.

[12] Collobert R, Bengio S, Bengio Y. A parallel mixture of SVMs for very large scale problems. Neural Computation 2002; 5: 1105 – 1114.

[13] Lu Y, Roychowdhury V, Vandenberghe L. Distributed Parallel Support Vector Machines in Strongly Connected Networks. IEEE Transactions on Neural Networks 2008; 7: pp. 1167 - 1178.

[14] Ruping S. Incremental Learning with Support Vector Machines. In: First IEEE International Conference on Data Mining (ICDM'01); 29 November – 2 December 2001; San Jose, California, USA: IEEE. pp. 641-642.

[15] Syed NA, Huan S, Kah L, Sung K. Incremental learning with support vector machines. In: Fifth ACM SIGKDD International Conference on Knowledge Discovery; 15-18 August 1999; San Diego, California, USA. ACM. pp. 317-321.

[16] Caragea C, Caragea D, Honavar V. Learning support vector machine classifiers from distributed data sources. In: Twentieth National Conference on Artificial Intelligence (AAAI); 9-13 July 2005; Pittsburgh, Pennsylvania, USA. AAAI. pp. 1602-1603.

[17] Sun Z, Fox G. Study on Parallel SVM Based on MapReduce. In: International Conference on Parallel and Distributed Processing Techniques and Applications; 16-19 July 2012; Las Vegas, USA. CSREA Publishing. pp. 495-561.

[18] Catak FO, Balaban ME. CloudSVM : Training an SVM Classifier in Cloud Computing Systems. In: Pervasive Computing & Networked World; 28-30 November 2012; Istanbul, Turkey. Springer. pp. 57-68.

[19] Frey PW, Slate DJ. Letter Recognition Using Holland-style Adaptive Classifiers. Machine Learning; 1991; 2: 161-182.

[20] Alimoglu F, Alpaydin E. Methods of combining multiple classifiers based on different representations for pen-based handwriting recognition. In: 4th International Conference on Document Analysis and Recognition; 1997; Washington, USA. IEEE. pp. 637-640.

[21] Stoyanov V, Ropson A, Eisner J. Empirical Risk Minimization of Graphical Model Parameters Given Approximate Inference, Decoding, and Model Structure. In: Fourteenth International Conference on Artificial Intelligence and Statistics AISTATS; 11-13 April 2011; Florida. JMLR. pp. 725-733.

[22] Vapnik V. An overview of statistical learning theory. IEEE Transactions on Neural Networks 1999; 10: 988-999.

[23]     Mercer J. Functions of positive and negative type and their connection with the theory of integral equations. Philosophical Transactions of the Royal Society 1909; 209: 415–446.

[24]     Dean J, Ghemawat S. Simplified data processing on large clusters. In: 6th conference on Symposium on Operating Systems Design & Implementation(OSDI); 6-8 December 2004; Berkeley, USA: ACM. pp. 107-113.

[25]     Schatz M. CloudBurst: Highly Sensitive Read Mapping with MapReduce. Bioinformatics 2009; 25: 1363-1369.

[26]     Rosasco L, Vito ED, Caponnetto A, Piana M, Verri A. Are loss functions all the same. Neural Computation 2011; 16: 1063-1076.